# Image Segmentation in Video Sequences: A Probabilistic Approach

**Nir Friedman, Stuart Russell**
Computer Science Division
University of California, Berkeley, CA 94720-1776
{nir,russell}@cs.berkeley.edu

## Abstract

"Background subtraction" is an old technique for finding moving objects in a video sequence—for example, cars driving on a freeway. The idea is that subtracting the current image from a time-averaged background image will leave only non-stationary objects. It is, however, a crude approximation to the task of classifying each pixel of the current image; it fails with slow-moving objects and does not distinguish shadows from moving objects. The basic idea of this paper is that we can classify each pixel using a model of how that pixel looks when it is part of different classes. We learn a mixture-of-Gaussians classification model for each pixel using an unsupervised technique—an efficient, incremental version of EM. Unlike the standard image-averaging approach, this automatically updates the mixture component for each class according to likelihood of membership; hence slow-moving objects are handled perfectly. Our approach also identifies and eliminates shadows much more effectively than other techniques such as thresholding. Application of this method as part of the Roadwatch traffic surveillance project is expected to result in significant improvements in vehicle identification and tracking.

## 1 Introduction

Finding moving objects in image sequences is one of the most important tasks in computer vision. For many years, the "obvious" approach has been first to compute the stationary *background image*, and then to identify the moving objects as those pixels in the image that differ significantly from the background. We will call this the *background subtraction* approach. The details of the method are described briefly in Section 2.

In earlier work as part of the Roadwatch project at Berkeley, it was shown that background subtraction can provide an effective means of locating and tracking moving vehicles in freeway traffic [Koller *et al.*, 1994]. Moving shadows do, however, cause serious problems, since they differ from the background image and are therefore identified as parts of the moving objects. Moreover, when traffic is slow-moving or stationary, the background image becomes corrupted by the vehicles themselves.

These problems arise from an oversimplified view of the task. What we would like to do is to classify each pixel of each image as moving object, shadow, or background. The basic idea of this paper is that we can classify each pixel using a probabilistic model of how that pixel looks when it is part of different classes (Section 3). For example, a given road pixel in shadow looks different from the same pixel in sunlight or as part of a vehicle. Because the appearance of the pixel in shadow is independent of the object that is casting the shadow, the shadow model for the pixel is relatively constant, like the background model. Furthermore, the probabilistic classification of the current pixel value can be used to update the models appropriately, so that vehicle pixels do not become mixed in with the background model when traffic is moving slowly.

Essentially, the pixel models describe the probability distribution of the appearance of the pixel conditioned on its type, where the type is a hidden variable. In this paper, the pixel appearance is modeled as a mixture of Gaussians, and is learned using the EM algorithm [Dempster *et al.*, 1977]. The details are given in Section 4, where we also describe an incremental version of the algorithm that provides real-time performance. We show that our approach is successful in coping with slow-moving objects and shadows (Section 5).

There is a large literature (several hundred papers in the last decade, and several annual conferences) on the application of EM and related techniques to image reconstruction, image segmentation, and motion identification. Applications include optical astronomy, laser range finders, synthetic aperture radar, MRI, PET, microscopy, and X-rays. Almost without exception, EM is used to identify classes of pixels within an image or classes of motions within an optical flow field, on the assumption that similar pixels can be grouped together. Typical examples include Samadani [1995], Jepson and Black [1993], Sawhney and Ayer [1996], and Weiss and Adelson [1996]. To our knowledge, the use of EM to model the appearance of a single pixel over time is novel, and provides a natural probabilistic generalization of a classical deterministic method.

## 2  Background subtraction

The roots of background subtraction go back to the 19th century, when it was shown that the background image could be obtained simply by exposing a film for a period of time much longer than the time required for moving objects to traverse the field of view. Thus, in its simplest form, the background image is the long-term average image:

$$B(x,y,t) = \frac{1}{t} \sum_{t'=1}^{t} I(x,y,t')$$

where $I(x, y, t)$ is the instantaneous pixel value for the $(x, y)$ pixel at time $t$. This can also be computed incrementally:

$$B(x,y,t) = \frac{(t-1)}{t} B(x,y,t-1) + \frac{1}{t} I(x,y,t)$$

The variance can also be computed incrementally, and moving objects can be identified by thresholding the Mahalanobis distance between $I(x, y, t)$ and $B(x, y, t)$.

One obvious problem with this approach is that lighting conditions change over time. This can be handled using a moving-window average, or, more efficiently, using exponential forgetting. In the latter scheme, each image's contribution to the background image is weighted so as to decrease exponentially as it recedes into the past. This is implemented by the update equation

$$B(x,y,t) = (1-\alpha)B(x,y,t-1) + \alpha I(x,y,t) \qquad (1)$$

where $1/\alpha$ is the time constant of the forgetting process. Unlike the moving-window method, this requires no additional storage. Exponential forgetting is equivalent to using a Kalman filter to track the background image, as done in [Koller et al., 1994].

Figure 1 shows a typical result from the method operating under favourable conditions. Although there are a few stray pixels identified as "moving" due to image noise, the vehicles are outlined reasonably well. Standard methods can be used to group the pixels belonging to each vehicle and to compute and track a smoothed convex hull.

The sharp-eyed reader will have spotted that the background subtraction method succeeds not only in detecting moving vehicles, but also their shadows.[1] In practice, shadows have been one of the most serious problems for video-based traffic surveillance in both commercial and research systems [Michalopoulos, 1991], sometimes resulting in undercounting or overcounting by as much as 50%. It might be thought that some simple fix such as lightening or thresholding might work to eliminate shadows, but these schemes may fail because parts of the road may be shadowed by buildings, and because of road markings—a shadow falling on a white line can still result in a brighter pixel than sunlight falling on tarmac [Kilger, 1992].

---

[1] Some of the road markings are also labelled as "moving"—this is due to camera jitter. Also, the method fails to detect those parts of a moving vehicle that are approximately the same intensity as the background. Such problems are unavoidable in any pixel-based method.

As mentioned in the introduction, another serious problem arises when objects are slow-moving or temporarily stationary. Here, "slow-moving" means that the time of traversal is non-negligible compared with $1/\alpha$, the time constant of the exponential forgetting process in Equation (1). When this happens, the background image becomes corrupted and object detection fails completely (Figure 2).

The solution used by Koller et al. [1994] was to update the background image with only those pixels *not* identified as moving objects. This is reasonably effective, but still has problems with very slow traffic because erroneous pixel classifications perturb the background image, which increases the number of erroneous classifications, and so on. Essentially, the distribution of pixels erroneously classified as non-moving is biased away from the true mean of the background image, causing instability in the process.

Despite these problems, the idea behind this approach is essentially right: each pixel must be *classified* before being used to update the background model. The next section shows how this can be done properly, using a probabilistic classifier and a stable updating algorithm. This approach also solves the problem with shadows.

## 3  Pixel models

Consider a single pixel and the distribution of its values over time. Some of the time it will be in its "normal" background state—for example, a small area of the road surface. Some of the time it may be in the shadow of moving vehicles, and some of the time it may be part of a vehicle. Thus, in the case of traffic surveillance, we can think of the distribution of values $i_{x,y}$ of a pixel $(x, y)$ as the weighted sum of three distributions $r_{x,y}$ (road), $s_{x,y}$ (shadow), and $v_{x,y}$ (vehicle):

$$i_{x,y} = \mathbf{w}_{x,y} \cdot (r_{x,y},\ s_{x,y},\ v_{x,y})$$

These distributions are subscripted to emphasize that they differ from pixel to pixel; $r_{x,y}$ is a probability distribution for the way that *this specific pixel* looks when it is showing unshadowed road at the corresponding physical location. It is essential to have different models for each pixels, because, for example, some parts of the image may correspond to white road markings, others to dark streaks in the centers of lanes, and still others to lamp-posts (see Figure 1). The weights are also subscripted, because some pixels may spend more time in shadow or vehicle than others.

Figures 3(a) and 3(b) show the empirical distribution of intensity and RGB values, respectively, for pixel (160,170),

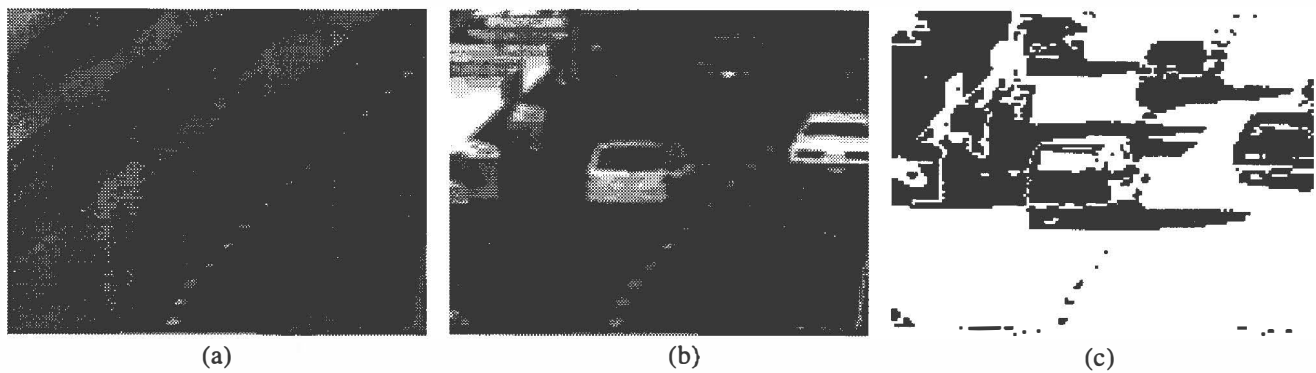

Figure 1: (a) Background image computed during fast-moving traffic using exponential forgetting. (b) Current image (frame 100). (c) Thresholded difference image showing pixels associated with moving objects.

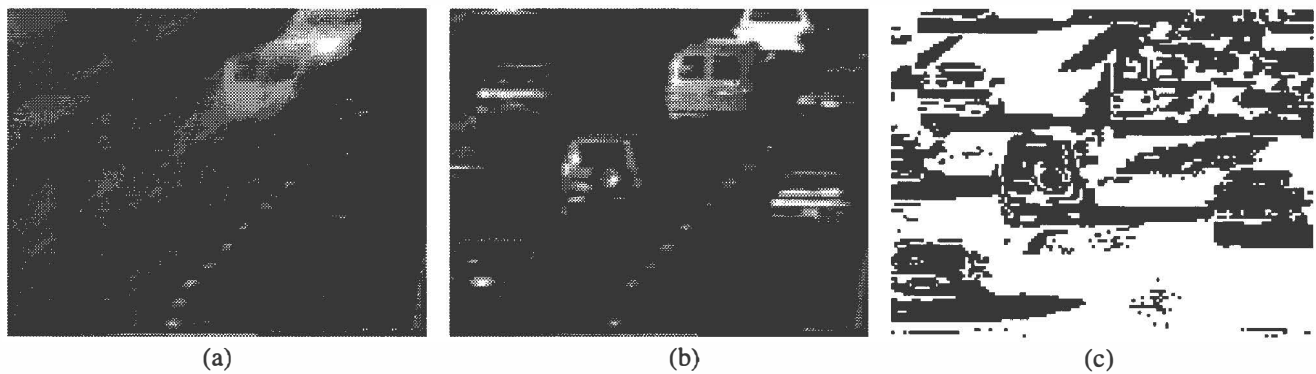

Figure 2: (a) Background image computed during slow-moving traffic using exponential forgetting. (b) Current image (frame 925). (c) Thresholded difference image showing pixels associated with moving objects.

which is roughly two-thirds of the way towards the bottom right corner of the image. These data display the behaviour one would expect: the shadow and road pixels form two fairly well-defined peaks, while the vehicle pixels are widely scattered. As a first approximation, we assume that each distribution can be modelled as a Gaussian. Using the techniques described in Section 4, we can fit three-component mixture models to the data. Figure 3(c) shows the fitted model for intensity values, and Figure 3(d) shows a scatter plot for the fitted RGB model. The fitted models are reasonably good (but far from ideal) approximations to the empirical data.

The model for pixel $(x, y)$ is parameterized by the parameters $\Theta = \{w_l, \mu_l, \Sigma_l : l \in \{r, s, v\}\}$ so that $\mathbf{w}_{x,y} = (w_r, w_s, w_v)$, $r_{x,y} \sim N(\mu_r, \Sigma_r)$, and so on.[2] Our models apply in two settings. In the first, we examine intensity levels, and $\mu$ and $\Sigma$ are scalars. In the second, we examine RGB values, and $\mu$ is a $3 \times 1$ vector and $\Sigma$ is a $3 \times 3$ matrix. The derivations are identical in the two cases, so we do not distinguish between them in the following discussion.

Let $i$ be a pixel value (either an intensity level or a vector of RGB values). Let $L$ be a random variable denoting the *label* of the pixel in this image. Our model defines the probability

---

[2]For clarity, we omit the subscript $x, y$ from the names of these parameters. However, it should be clear that there is a different set of parameters for pixel position $x, y$.

that $L = l$ and $I(x, y, t) = i$ to be

$$P(L = l, I(x, y, t) = i \mid \Theta) =$$

$$w_l \cdot (2\pi)^{-\frac{d}{2}} |\Sigma_l|^{-\frac{1}{2}} \exp\{-\frac{1}{2}(i - \mu_l)^T \Sigma_l^{-1}(i - \mu_l)\}$$

where $d$ is the dimension of each pixel value (1 or 3 in our case). Given these probabilities, we can classify the pixel value. Namely, we choose the class $l$ with highest posteriori probability $P(L = l \mid I(x, y, t))$.

## 4 Algorithms for learning pixel models

This section begins by describing the standard EM algorithm for learning mixture models from observed data with the class variable hidden. We then describe an efficient, incremental version suitable for real-time implementation.

### 4.1 EM for mixture models

Suppose we observe a sequence of pictures $1, \ldots, T$, and that $I(x, y, t)$ is the value of pixel $(x, y)$ in the $t$th image. We want to learn the parameters of the distributions $r_{x,y}, s_{x,y}, v_{x,y}$ as well as the relative weights $\mathbf{w}_{x,y}$. To formally set up the learning problem, we define the *likelihood* of a set of parameters $\Theta$ to be the probability of the data given $\Theta$: $\prod_{t=1}^{T} P(L = l_t, I(x, y, t) \mid \Theta)$. We want to choose the parameters that maximize the likelihood.

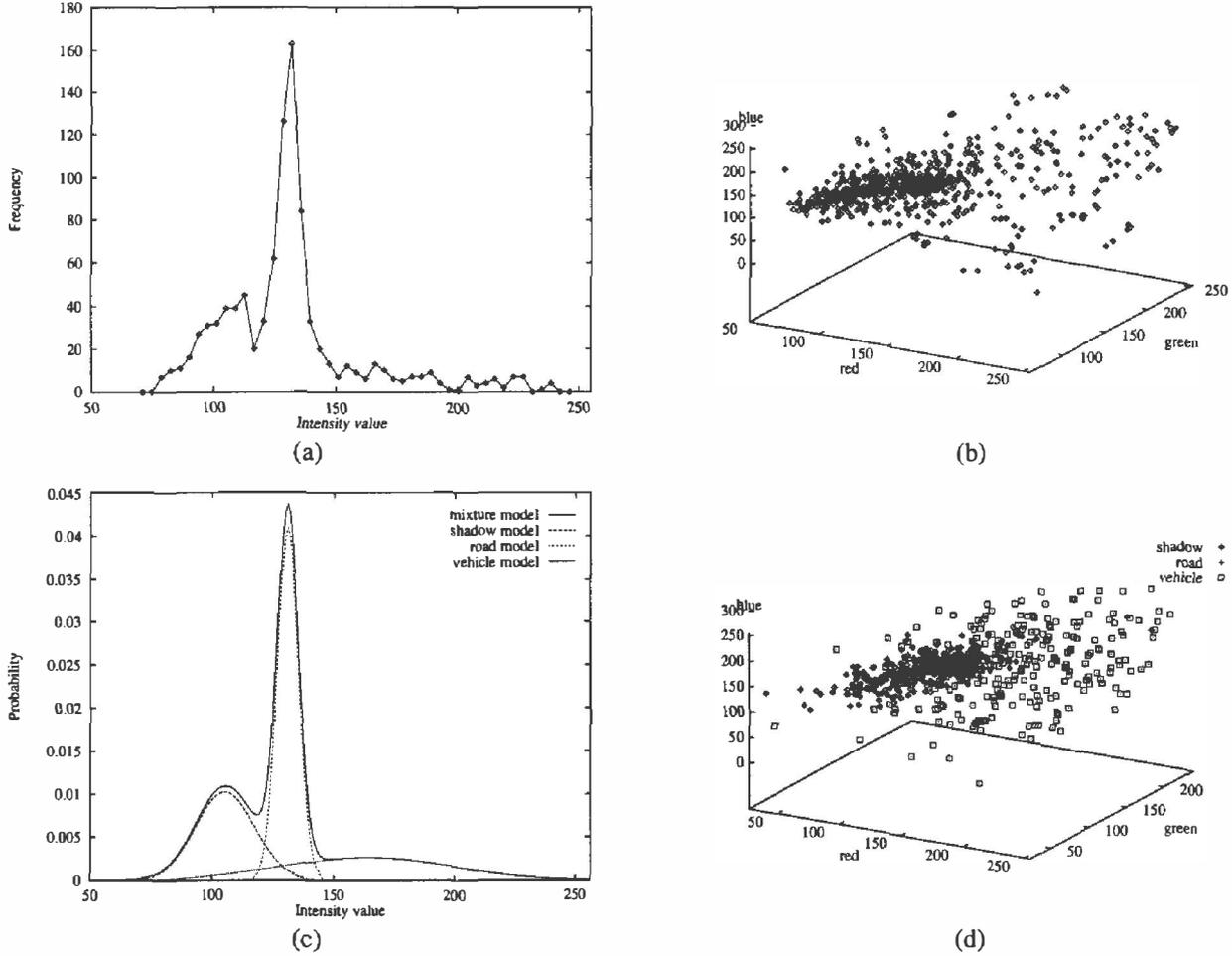

Figure 3: (a) Empirical distribution of intensity values for pixel (160,170) over 1000 frames. (b) Scatter plot of RGB values for the same pixel. (c) Fitted three-component Gaussian mixture model for the data in (a). (d) Scatter plot of 1000 randomly-generated data points from a fitted three-component Gaussian mixture model for the data in (b).

We begin by examining a simpler problem. Suppose that the images were annotated by the labels of all pixels, and suppose that $L_t(x, y)$ is the label of $x,y$ in the $t$th image. In this case, learning these parameters would be easy. Standard arguments show that the optimal settings of parameters for this case can be computed as follows. We define the *sufficient statistics* for this mixture estimation to be $N_l$, $M_l$, and $Z_l$, where

- $N_l(x, y)$ is the number of images for which $L_t(x, y) = l$;
- $M_l(x, y)$ is the sum of the pixel values for which $L_t(x, y) = l$, which we write as $\sum_{t=1,...,T, L_t=l} I(x, y, t)$; and
- $Z_l(x, y)$ is given by $\frac{1}{N_l(x,y)} \sum_{t=1,...,T, L_t=l} I(x, y, t) \cdot I(x, y, t)^T$, the sum of the outer products of the input vectors with themselves.

From these sufficient statistics, we can compute $w_l$, $\mu_l$, and $\Sigma_l$ as:

$$w_l = \frac{N_l}{\sum_{l'} N_{l'}} \quad (2)$$

$$\mu_l = \frac{M_l}{N_l} \quad (3)$$

$$\Sigma_l = \frac{1}{N_l} Z_l - \mu_l^T \mu_l \quad (4)$$

Unfortunately, we do not have labels $L_t$ for our training data. Thus, we define the likelihood with respect to the observable data: $L(\Theta) = \prod_{t=1}^{T} P(I(x, y, t) \mid \Theta)$. Learning mixture models is one of the classic examples of *missing values*. The standard solution in the literature is the *expectation maximization* algorithm [Dempster et al., 1977; McLachlan and Krishnan, 1997]. Roughly speaking, the EM algorithm explores a sequence of parameter settings, where each setting is found by using the previous one to classify the data. More precisely, assume that we have some parameter setting $\Theta^k$. We can use the probability of different labels according to $\Theta^k$ as an estimate of their true distribution. We now reestimate the parameters, where we "split" each vector $I(x, y, t)$ between the different Gaussians according to these proportions. Formally, we compute the *expected value* of the sufficient statistics as follows:

$$E[N_l \mid \Theta^k] = \sum_{t=1}^{T} P(L_t = l \mid I(x, y, t), \Theta^k)$$

$$E[M_l \mid \Theta^k] = \sum_{t=1}^{T} P(L_t = l \mid I(x, y, t), \Theta^k) I(x, y, t)$$

$$E[Z_l \mid \Theta^k] = \sum_{t=1}^{T} P(L_t = l \mid I(x, y, t), \Theta^k) \\ I(x, y, t) \cdot I(x, y, t)^T$$

We then define $\Theta^{k+1}$ by using Equations 2–4 with the expected sufficient statistics.

This process has two important properties. First, $L(\Theta^{k+1}) \geq L(\Theta^k)$. That is, $\Theta^{k+1}$ provides a better approximation to the distribution of the data. Second, if $\Theta^{k+1} = \Theta^k$, then $\Theta^k$ is a stationary point (e.g., a local maximum) of the likelihood function. Combining these two properties, we see that this procedure will eventually converge to a stationary point [McLachlan and Krishnan, 1997]. If we start the process several times with different setting of $\Theta^0$, we hope to find a good approximation to the optimal setting. In addition, for our application, we must have a way to identify which component of the model should be labelled as road, which as shadow, and which as vehicle. A heuristic solution to this problem is described in Section 5

### 4.2 Incremental EM

The standard EM procedure we just reviewed requires us to store the values of pixel $(x, y)$ for all the images we have observed. This is clearly impractical for our application. Moreover, batch processing of the complete image sequence is not possible in a real-time setting. We now describe an incremental variant of EM that does not require storing the data. This procedure was introduced by Nowlan [1991], and is best understood in terms of the results of Neal and Hinton [1993].

Neal and Hinton show that we can think of the EM process as continually adjusting the sufficient statistics. In this view, on each iteration when we process an instance, we remove its previous contribution to the sum and replace it with a new one. Thus, for example, when we update $N_l$, we remove $P(L_t = l \mid I(x, y, t), \Theta^{k'})$ and add $P(L_t = l \mid I(x, y, t), \Theta^k)$, where $\Theta^{k'}$ are the parameter settings we used to compute the previous estimated statistics from $I(x, y, t)$, and $\Theta^k$ are the current parameter settings. Similarly, to update $M_l$, we remove $P(L_t = l \mid I(x, y, t), \Theta^{k'}) I(x, y, t)$ and add $P(L_t = l \mid I(x, y, t), \Theta^k) I(x, y, t)$. Neal and Hinton show that after each instance is processed, the likelihood of the data increases.

This argument motivates the following incremental approach. Whenever we observe a new instance, we add its contribution to the sufficient statistics. This means that we are increasing our training set at each step, yet we never reprocess the previous instances in the training set. This procedure is no longer guaranteed to be monotonic in $L$, but *on the average* this process increases $L$. Thus, in the long run, this process converges to a local maximum with high probability.

The resulting procedure for each pixel $(x, y)$ has the following structure:

Initialize parameters $\Theta$.
$t \leftarrow 0$
for $l \in \{r, s, v\}$
   $N_l \leftarrow k w_l$
   $M_l \leftarrow k w_l \cdot \mu_l$
   $Z_l \leftarrow k w_l \cdot (\Sigma_l + \mu_l \cdot \mu_l^T)$
do forever
   $t \leftarrow t + 1$
   for $l \in \{r, s, v\}$
     $N_L \leftarrow N_L + P(L_t = l \mid I(x, y, t), \Theta)$
     $M_l \leftarrow M_l + P(L_t = l \mid I(x, y, t), \Theta) I(x, y, t)$
     $Z_l \leftarrow Z_l + P(L_t = l \mid I(x, y, t), \Theta) I(x, y, t) I(x, y, t)^T$
   Compute $\Theta$ from $\{N_L, M_l, Z_l\}$.

The initialization step of the procedure sets the statistics to be the expected statistics for the initial choice of $\Theta$. Then, in each iteration we add the expected statistics for the new instance to the accumulated statistics.

This procedure performs quite well, and the reported experiments in Section 5 are based on it. However, since the sufficient statistics terms keep growing, this procedure can run into problems in the long run. Moreover, the procedure never removes the contributions of old instances from the sufficient statistics. Intuitively, the models that were used to compute the expectation for these instances is quite out of date. Thus, the procedure would perform better if these instances were removed. To deal with both problems, we can introduce exponential forgetting, as done in Section 2. A version of Equation (1) can be applied to the incremental EM process quite straightforwardly. We are in the process of experimenting with this variant.

## 5 Empirical results

Our general procedure for processing a video sequence is as follows:

1. Initialize mixture models for each pixel with a weak prior;

2. For each new frame:

    (a) Update the estimated mixture model for each pixel using incremental EM;

    (b) Heuristically label the mixture components;

    (c) Classify each pixel according the its current mixture model.

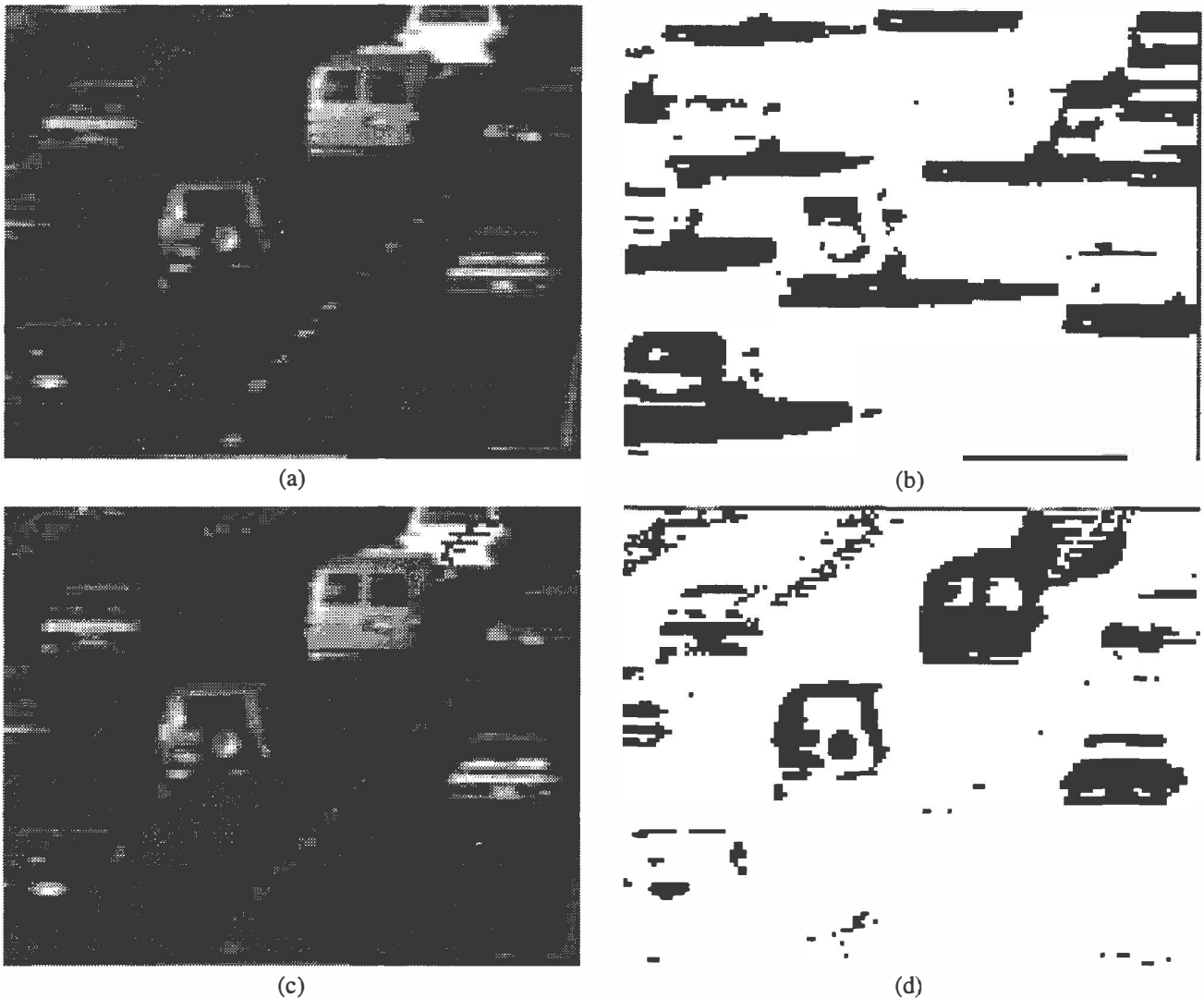

Figure 4: (a) Original image (frame 925). (b) Pixels identified as shadow. (c) Image with shadow pixels replaced by corresponding road value. (d) Mask showing pixels classified as vehicles.

The heuristic labelling process is needed in cases where the mixture components are not in the same order as the prior model indicates. For example, the prior model expects that vehicle pixels will be brighter, in general, than road pixels. For white road markings, the brightness order is reversed. Our heuristics are as follows: label the darkest component as shadow; of the remaining two components, label the one with the largest variance as vehicle and the other as road.

Sample results are shown in Figure 4, for the same image as shown in Figure 2. We show the original image and the identified shadow pixels, together with an image resulting from replacing shadow pixels with the corresponding road value. The mask image showing the vehicle pixels is much cleaner than that obtained by background subtraction. At the time of writing, we are beginning the experiments needed to show that this improvement carries over into vehicle detection and tracking performance. We are also rerunning our experiments with RGB models instead of intensity models; we assume that use of colour information will largely eliminate the tendency of darker vehicles to disappear, since very few vehicles have the same hue as the road, even if they have the same intensity values. We note that the incremental version of EM yields an algorithm capable of running in real time on a suitable platform. Our target hardware platform, to which we hope to port the algorithm, consists of 12 Texas Instruments C40 DSP chips running in parallel.

## 6  Conclusions and further work

We have shown that a probabilistic approach to pixel classification, together with an unsupervised learning algorithm for pixel models, can significantly improve the detection of moving objects in video sequences. This is just one sample point in a vast space of possible research on the application of probabilistic inference to the task of understanding sensory input, almost all of which is going on outside the AI community.

Several improvements are needed in our system before it can be fielded. The most important from the point of view of robustness is the need for better initialization and labelling of models. Our heuristic approach may not work well in extremes of lighting conditions. Currently, our weak prior is equivalent to an initial model with a certain amount of evidential support—we are not able, for example, to state that the vehicle component is expected to have large variance but that its mean is very uncertain. Such priors require a MAP or penalty-based version of EM. This should require little or no additional computational expense and may result in significantly more robust performance.

We expect to reach the limits of purely pixel-based techniques fairly soon. Using more background knowledge, encoded as probabilistic models, one can expect much better performance. For example, temporal and spatial contiguity are strong constraints that we currently ignore—or rather, they are enforced in the grouping code, which is entirely *ad hoc*. Temporal contiguity in a pixel's classification can be enforced using a simple Markov model in which any given classification has a high probability of persisting. Similar schemes can be used for spatial contiguity, but these require Markov networks that impose a high computational burden. In any case, this level of models should allow extensions to, e.g., moving cameras, and should interface nicely with the higher-level dynamic belief network models of vehicle behaviour used in [Huang *et al.*, 1994] to predict motions and to detect events such as stalled vehicles and accidents.